\pdfoutput=1

\documentclass[11pt]{article}

\usepackage[preprint]{acl}

\usepackage{lipsum}  
\usepackage{etoolbox}
\AtBeginEnvironment{quote}{\singlespacing\small}
\usepackage{amsmath}
\usepackage{multirow}

\usepackage{times}
\usepackage{latexsym}

\usepackage[T1]{fontenc}

\usepackage[utf8]{inputenc}

\usepackage{microtype}

\usepackage{inconsolata}

\usepackage{graphicx}

\title{Are LLMs Models of Distributional Semantics?\\ A Case Study on Quantifiers}

\makeatletter
\renewcommand*{\@fnsymbol}[1]{\ifcase#1\or $\:\lambda$ \else\@ctrerr\fi}
\makeatother

\author{Zhang Enyan\thanks{Equal contribution in alphabetical order.} \\
  Dept. of Computer Science \\
  Yale University \\
  \texttt{enyan.zhang@yale.edu}\\\And
  Zewei Wang$^{\:\lambda}$ \\ 
  Dept. of Computer Science \\
  Columbia University \\
  \texttt{altria.wang@columbia.edu}\\\And
  Michael A. Lepori \\
  Dept. of Computer Science \\
  Brown University \\
  \texttt{michael\_lepori@brown.edu}\\\AND
  Ellie Pavlick \\
  Dept. of Computer Science \\
  Brown University \\
  \texttt{ellie\_pavlick@brown.edu}\\\And
  Helena Aparicio \\
  Department of Linguistics \\
  Cornell University \\
  \texttt{haparicio@cornell.edu}
  } 

\begin{document}
\maketitle

\begin{abstract}
Distributional semantics is the linguistic theory that a word's meaning can be derived from its distribution in natural language (i.e., its use). Language models are commonly viewed as an implementation of distributional semantics, as they are optimized to capture the statistical features of natural language. It is often argued that distributional semantics models should excel at capturing graded/vague meaning based on linguistic conventions, but struggle with truth-conditional reasoning and symbolic processing. We evaluate this claim with a case study on vague (e.g. ``many'') and exact (e.g. ``more than half'') quantifiers. Contrary to expectations, we find that, across a broad range of models of various types, LLMs align more closely with human judgements on exact quantifiers versus vague ones. These findings call for a re-evaluation of the assumptions underpinning what distributional semantics models are, as well as what they can capture.
\end{abstract}

\section{Introduction}

Distributional semantics --- the thesis that word meaning can be derived from the distribution of a word's use --- is arguably the linchpin of contemporary NLP. For more than thirty years \cite{ir1990}, researchers and engineers have relied on it to build increasingly advanced language technologies by increasing both the quantity and diversity of training data \cite{kaplan2020scalinglawsneurallanguage} and by invoking more advanced methods for representing and predicting, rather than counting, contexts \cite{cbow, NIPS2013_9aa42b31, devlin-etal-2019-bert, baroni-etal-2014-dont}. Large language models (LLMs) appear to be the natural product of this intellectual and technical tradition.

In linguistic theory, distributional semantics is often held in direct contrast with formal semantics, i.e., semantics in the symbolic, logical tradition of Frege \cite{sep-frege-logic} and \citet{montague1974proper}. The conventional wisdom is that distributional semantics and formal semantics have complementary strengths, with the former excelling at modeling lexical semantics, in particular in cases that require context-sensitivity and gradedness \cite{gamma_review, wittgenstein2009philosophical}, and the latter excelling in compositional semantics, in particular in cases that require knowledge of truth conditions \cite{BeltageyMeaningCombinationof, coecke2010mathematicalfoundationscompositionaldistributional}. This dichotomy is widely accepted, and is the basis of many research programs focused on hybrid formal-distributional (i.e. neuro-symbolic) systems that can capture the best of both worlds \cite{baroni-etal-2014-frege, nye2020learning, drozdov2022compositional}.

The above intuition about the strengths of distributional semantics makes a straightforward prediction about the inferential capabilities of LLMs. Specifically, as distributional semantics models, LLMs should excel on inferences that exhibit gradedness and are derived from linguistic convention, and should falter on inferences which exhibit exactness and are derived from truth conditions. We test this prediction using quantifiers as a case study, with graded quantifiers (``many'', ``few'') representing the former type of inference and exact quantifiers (``more than half'', ``less than half'') the latter. We collect data from human subjects and compare it to inferences made by both open source and proprietary state-of-the-art LLMs. We find that, counterintuitively, the opposite trend holds: LLMs tend to match human inferences in cases involving exact quantifiers but struggle in cases involving vague quantifiers.

These results raise important questions about what it means to be a distributional semantics model. Either we must revisit our assumptions about what distributional semantics can capture, or we must determine at what point a model trained only on the distribution of text ceases to embody the distributional hypothesis. Either conclusion would carry important implications for semantic theory, and for the role of computational models for both building and testing such theory. 

\section{Methodology and Experiments}

\subsection{Background}

Quantifiers are lexical items that encode the (approximate) quantity of related nouns.  Here, we focus on quantifiers that express proportions between two cardinalities of objects (e.g., ``many'', ``few'', ``more/less than half'', ``all'' or ``none'', among others; Cf. \citet{partee-many}). In the formal semantics tradition, the meaning of proportional quantifiers is often stated as a relation between the relevant ratio and some threshold $\theta$ (Equation \ref{eq:quantifiers}), Cf. \citet{hackl2009}. For instance, the meaning of `Many A's are B's' is \texttt{true} on its proportional reading if and only if the number of $A$'s that are $B$'s constitutes a proportion of $A$'s that is equal or higher than some threshold $\theta$. 
\begin{equation}\label{eq:quantifiers}
   [\![\text{Many} ]\!](A)(B) = \texttt{true} \;\;\text{iff}\;\; \frac{|A \cap B|}{|A|}\geq \theta
\end{equation}

For exact quantifiers, the value of the threshold variable is specified in the semantics (e.g., for the quantifier ``all'', the ratio must be equal to 100\%, whereas for ``more than half'' the ratio must be higher than 50\%). However, for vague quantifiers, such as ``many'' and ``few'', the threshold variable is semantically \emph{underspecified} and must be approximated in conversation. Importantly, this approximation process is highly context-sensitive: listeners rely on different sources of information to infer plausible threshold values for vague quantifiers, such as world knowledge (e.g., the nature of the objects under discussion), previous linguistic exposure, or the individual production preferences of other speakers \cite{10.3389/fpsyg.2015.00441, pecsok2024faultless}. As a consequence, the meaning of vague quantifiers, as opposed to that of exact quantifiers, can be highly variable.

\subsection{Experimental Design}

Our experiment tests judgements on proportional quantifiers ``many", ``few", ``more than half" and ``less than half". We characterize the semantics of these quantifiers by measuring the threshold empirically. For both humans and language models, we ask a question in the following form:

\texttt{There are <total-number> <noun>. <number> of them are yellow. Are many <noun> yellow?}

To mitigate confounding factors from nouns (for example, ``grains of sand'' may have a different threshold than ``cars''), we select a range of objects for the \verb|<noun>| position: \textit{bicycles, tables, balls, books, pens, cars, phones, cups, clocks, computers}. We also vary the \verb|<total-number>| by choosing from a set of randomly generated numbers from 10 to 1000 to increase result robustness and test the stability of thresholds. At test time, we ask for a yes/no judgement with regards to a specific, randomly sampled \verb|<number>| within the range of $[0, \verb|total_number|]$. The threshold for a quantifier is then inferred from the responses to these questions. 

Our prompt design encourages a proportional interpretation of the quantifier by explicitly mentioning a relevant \verb|<total-number>|. This has two main advantages: as ``more/less than half'' only elicit proportional readings, it creates a closer parallel between the two sets of quantifiers. Furthermore, this design also controls for task complexity: unlike ``all/none'', which can be solved as a comparison task without invoking a notion of threshold ($x$ = \verb|<total-number>| and $x = 0$, respectively), judging both ``few/many'' and ``less/more than half'' requires making proportional comparisons between two numbers. It helps mitigate the possibility that performance is better on a set of quantifiers purely because the task is simpler, especially for language models. In the following sections, we outline our procedures for collecting data from humans and language models.

\subsection{Human Experiments}\label{sec:human_section}
 
Each quantifier (i.e., ``few'', ``many'', ``more than half'', ``less than half'') was tested across four parallel experiments, where each set of experiment includes a total of 1000 questions. Collectively, these questions cover a combination of 25 \verb|<total-number>|'s, 20 \verb|<number>|'s for each \verb|<total-number>|, spread across 10 \verb|<noun>|'s. The 25 \verb|<total-number>| were chosen randomly between 10 and 1000, and for each \verb|<total-number>|, 20 \verb|<number>|'s were randomly sampled between the range of $[0, \texttt{<total-number>}]$. To ensure that target \verb|<number>|'s are distributed in a way such that the responses to these questions will be  informative about the threshold, the 20 \verb|<number>|'s are sampled in the following way:  the \verb|<number>|'s were above 50\% of its respective \verb|<total-number>|, and half were below; furthermore, half of the \verb|<number>|'s were round numbers and half were non-round\footnote{For the \texttt{<total-number>} 1000, round numbers are defined as multiples of 50. For all smaller \texttt{<total-number>}'s, round numbers are defined as multiples of 5.}The detailed procedure for composing questions as well as specific numbers used are presented in Appendix \ref{appendix:exp_humans}.

The set of 1000 total questions is then divided into 20 groups, each containing 50 questions. Each group's questions shares two \verb|<noun>|'s, but each of the questions uses a different \verb|<total-number>| - \verb|<number>| pair. Each participant was assigned to one such group, with questions presented in random order. Finally, each group contains an additional 2 attention-check trials. Attention trials includes only exact quantifiers from the set \{``all'', ``at least one''\}, which all have a easy-to-solve ground truth answer, and tested a different set of \verb|<noun>|'s than those used in experimental trials. Prior to the start of the experiment, each participant also receives 2 practice trials to help them familiarize with the experimental task, where practice trials are designed in the same way as attention-check trials.

As participants' task was to answer the question with their keyboard, an additional instruction for how to provide responses was appended. See below for an example question presented to the participant:

\vspace{0.5em}
\noindent
\texttt{
Question: There are 500 balls. 234 of them are yellow. Are many balls yellow? If yes, please press ‘J’ on your keyboard, otherwise press ‘F’.   
}
\vspace{0.5em}

Participants did not have a time limit to provide their answers. Explicit instructions were given to not use a calculator. For each set of experiment, we recruited 100 participants (5 per group of questions) using the crowdsourcing web platform \emph{Prolific} (\url{https://www.prolific.com/}). All participants were native speakers of English above 18 years old. e filtered out 38 out of 400 participants who did not perform perfectly on attention trials. 

\subsection{Language Model Experiments}

We query the same set of questions presented to human test participants on a range of state-of-the-art language models of varying sizes and types: InternLM-2.5 8b and its instruction-tuned version \cite{cai2024internlm2technicalreport}, Llama-3 8b, 70b, and respective instruction-tuned versions \cite{dubey2024llama3herdmodels}, Qwen-2.5 32b, 72b, and respective instruction-tuned versions \cite{yang2024qwen2technicalreport}, Mistral 7b and its instruction-tuned version \cite{jiang2023mistral7b}. For closed-source models served over APIs, we query the de-facto standard frontier models, GPT-3.5-Turbo, GPT-4-Turbo, and GPT-4o\footnote{The specific versions of the models are \texttt{gpt-3.5-turbo-0125}, \texttt{gpt-4-turbo-2024-04-09}, and \texttt{gpt-4o-2024-05-13}, which are the newest versions at the time of the experiment.}. We query all models with their default, recommended inference settings.


As we hope to have a parallel setting for instruction-tuned and pretrained base models, we append an instruction for answering the question that is compatible with both model types. An example question presented to models is shown below:

\vspace{0.5em}
\texttt{
Question: There are 500 balls. 234 of them are yellow. Are many balls yellow? Choices: Yes or No. \\ Answer:   
}
\vspace{0.5em}

In practice, this prompt works well for all models queried, as the token output with largest probability is always either ``yes'' or ``no''. This indicates that all models follow the basic instructions of the prompt and responds with a yes/no choice, allowing us to compare their outputs to human data, which is based on the same yes/no judgement. 

In addition, language models are known to be sensitive to the wordings of their prompts \cite{webson-pavlick-2022-prompt, zhan2024unveilinglexicalsensitivityllms}. To mitigate this issue, we test a total of 5 semantically equivalent sentence structures in place of the question ``Are 
\verb|<quantifier>| \verb|<noun>| yellow?", and report the average performance across prompts. Appendix \ref{appendix:exp_lms} details the experimental design. 

\section{Quantifier Thresholds in Humans}

\subsection{Data Preprocessing}

We first filter out responses provided by respondents that failed attention checks\footnote{Attention checks are questions involving a simple quantifier with a ground truth answer, e.g.,  ``Would you say that \emph{at least one} \texttt{<noun>} is yellow?'' instead. We assume that all attentive respondents will be able to perform the simple comparison task correctly.}. We then aggregate the remaining responses by collapsing over different \verb|<noun>|'s (i.e., if a set of questions contains the same \texttt{(<total-number>, <number>, <quantifier>)} tuple, we consider them to be the same set). Assuming that the connotations of different \verb|<noun>|'s only introduce confounding factors that are external to the semantics of the quantifier, aggregated results from this set of questions should reflect human judgements of the quantifier within the context the supplied numbers. We define the \textit{proportion} of this set of questions as $\frac{\texttt{<number>}}{\texttt{<total-number>}}$, and \textit{agreement} as $\frac{|Y|}{|Y| + |N|}$, where $Y$ and $N$ are the sets of question-answer pairs for which the response was ``Yes'' and ``No'', respectively.

\subsection{Human Thresholds}

\begin{figure}
    \centering
    \includegraphics[width=1.05\linewidth]{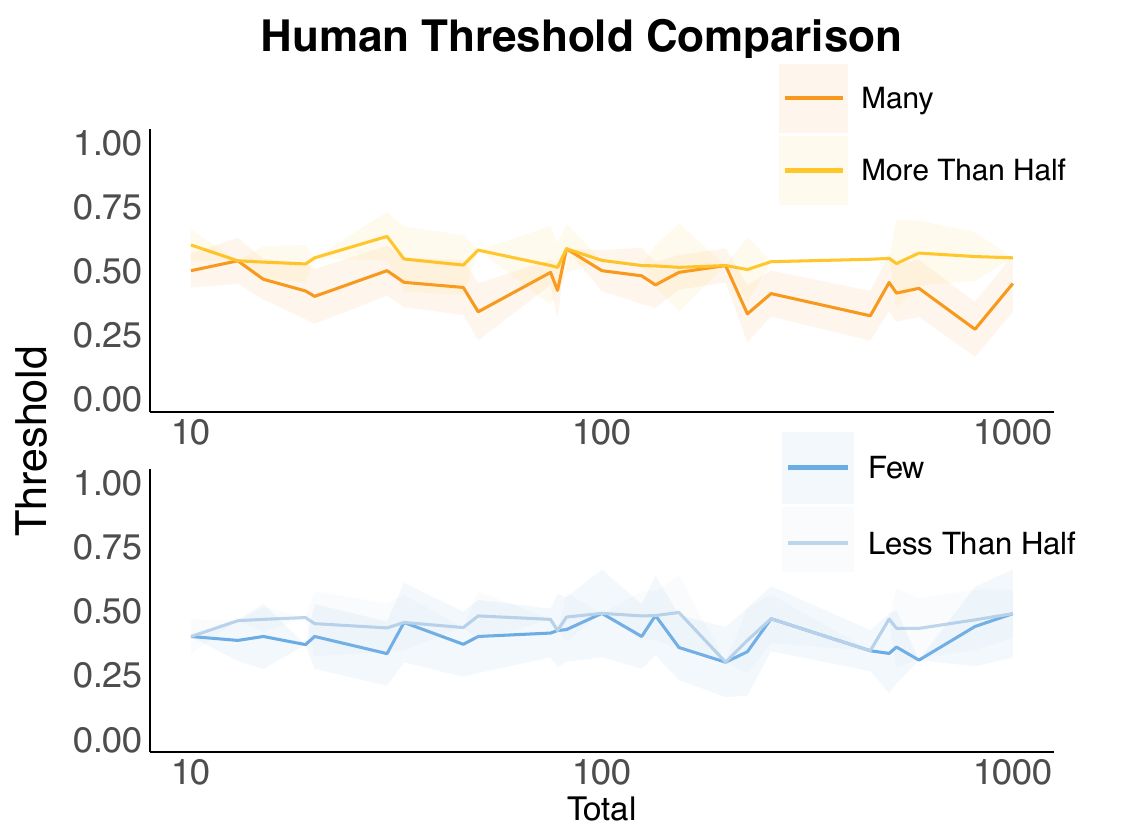}
    \caption{Human judgment of proportional threshold plotted as a function of total number; ribbons represent standard deviation. The horizontal axis is on log-scale. Human judgements for the threshold of vague and exact quantifiers are similar, and are roughly constant across different total numbers.}
    \label{fig:human}
\end{figure}

After preprocessing we obtain a set of (\textit{proportion}, \textit{agreement}) pairs for each \texttt{<total-number>}. We infer the threshold as the critical proportion where agreement crosses $50\%$. For quantifiers where the agreement is expected to correlate positively with proportion (i.e. a higher $\frac{\texttt{<number>}}{\texttt{<total-number>}}$ is more likely to satisfy the truth conditions of the quantifier, such as ``many'') this would mean the lowest proportion where agreement $\geq 51\%$. For quantifiers where agreement is expected to correlate negatively with proportion (such as ``few'') this would mean the highest proportion where agreement $ \leq 49\%$. Intuitively, this quantity reflects the lowest proportion which the majority of the participants agree is ``many''. 

Even though there is an explicit threshold of ``more/less than half'' of 50\%, the design of the experiment required the participants to conduct a rough estimate of the proportion based on actual provided numbers, which accounts for their real-world performance (as opposed to idealized competence, see \citet{ChomskyAspectsTheorySyntax}). We observe for both exact quantifiers a relatively steady and flat threshold across all \texttt{<total-number>}'s in Figure \ref{fig:human}. Aggregating data from all \verb|<total-number>|'s shown in Figure \ref{fig:human} gives the estimated threshold for each quantifier: $\theta_{more\ than\ half} = 0.547$, $\sigma_{more\ than\ half} = 0.034$, $\theta_{less\ than\ half} = 0.446, \sigma_{less\ than\ half} = 0.067$. 

The thresholds of ``many" and ``few" across different \verb|<total-number>|'s also remain relatively constant. We estimate the threshold from all trials as $\theta_{many} = 0.444$, $\sigma_{many} = 0.123$, $\theta_{few} = 0.395, \sigma_{few} = 0.118$. We observe that the standard deviation of vague quantifier thresholds are overall higher than those of exact quantifiers, which is consistent with their gradedness and context sensitiveness.

To verify whether the threshold is independent from their respective \verb|<total-number>|'s, we fit a linear regression threshold as a function of $\log$ (\verb|<total-number>|) . The slope terms were insignificant (at $p = 0.05$) for all quantifiers (``many": $\beta= -0.033$, $p= 0.157$; ``few": $\beta= -0.003$, $p= 0.859$; ``more than half": $\beta= -0.009$, $p= 0.417$; ``less than half": $\beta=-0.008$, $p= 0.640$). 

\section{Quantifier Thresholds in Language Models} 
\label{sec:results-lm}

For language models, we calculate threshold with the same process, except by using $\frac{P(\text{Yes})}{P(\text{Yes}) + P(\text{No})}$ instead of $\frac{|Y|}{|Y| + |N|}$. Each language model will have a threshold corresponding to each set of prompts, similar to human thresholds. Assuming that prompt choices are irrelevant to the semantics of quantifiers, we aggregate results across different prompts when reporting performance.

\begin{figure*}[ht]
\centering
\includegraphics[width=\linewidth]{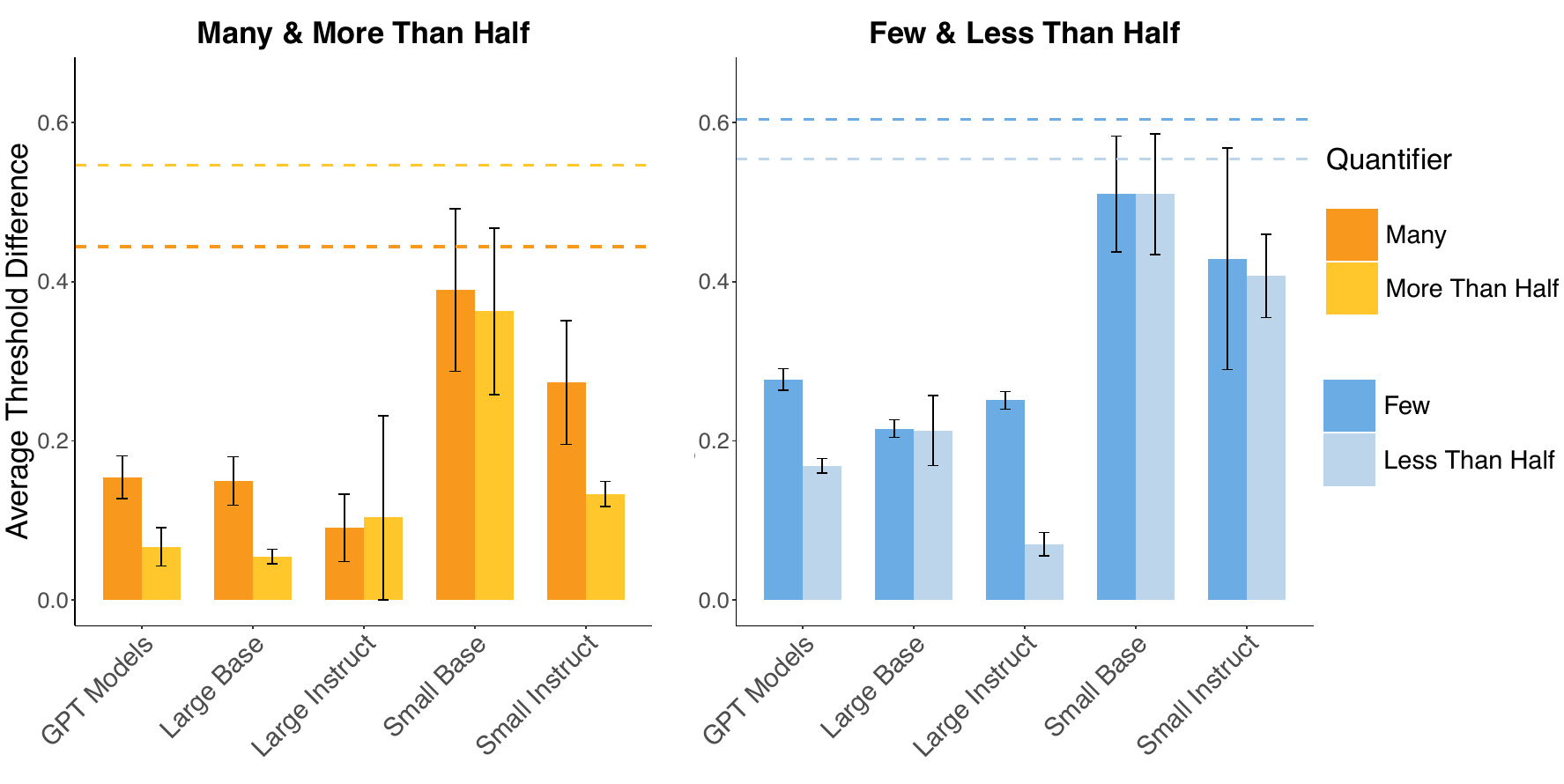}
\caption{Performance comparison between exact quantifiers and vague quantifiers, where each bar represents difference in threshold with human judgements (smaller is better); Error cap (dotted line) represents the maximum difference possible compared with human (i.e., responding yes or no to all prompts); Error bars represent standard deviation averaged across models within the group. To the extent that difference between vague/exact quantifiers is significant, LLMs always perform better on exact quantifiers compared to vague quantifiers.}
\label{fig:avg_difference}
\end{figure*}

\subsection*{Finding 1: LLMs are better models of human judgements for exact quantifiers than vague quantifiers}

We obtain a metric for how well model responses reflect human judgements by calculating the absolute difference (i.e., $|\theta_{\text{human}} - \theta_{\text{model}}|$) between the threshold of a model and human participants, and averaging them across all prompts. These are reported as ``Average Threshold Difference'' in Figure \ref{fig:avg_difference}. For example, if humans judge that, on average, the minimum proportion required for meeting the conditions of ``many'' is 45\%, and a model judges 35\%, the difference would be 10\%.

\begin{figure*}[ht!]
\centering
\includegraphics[width=0.95\linewidth]{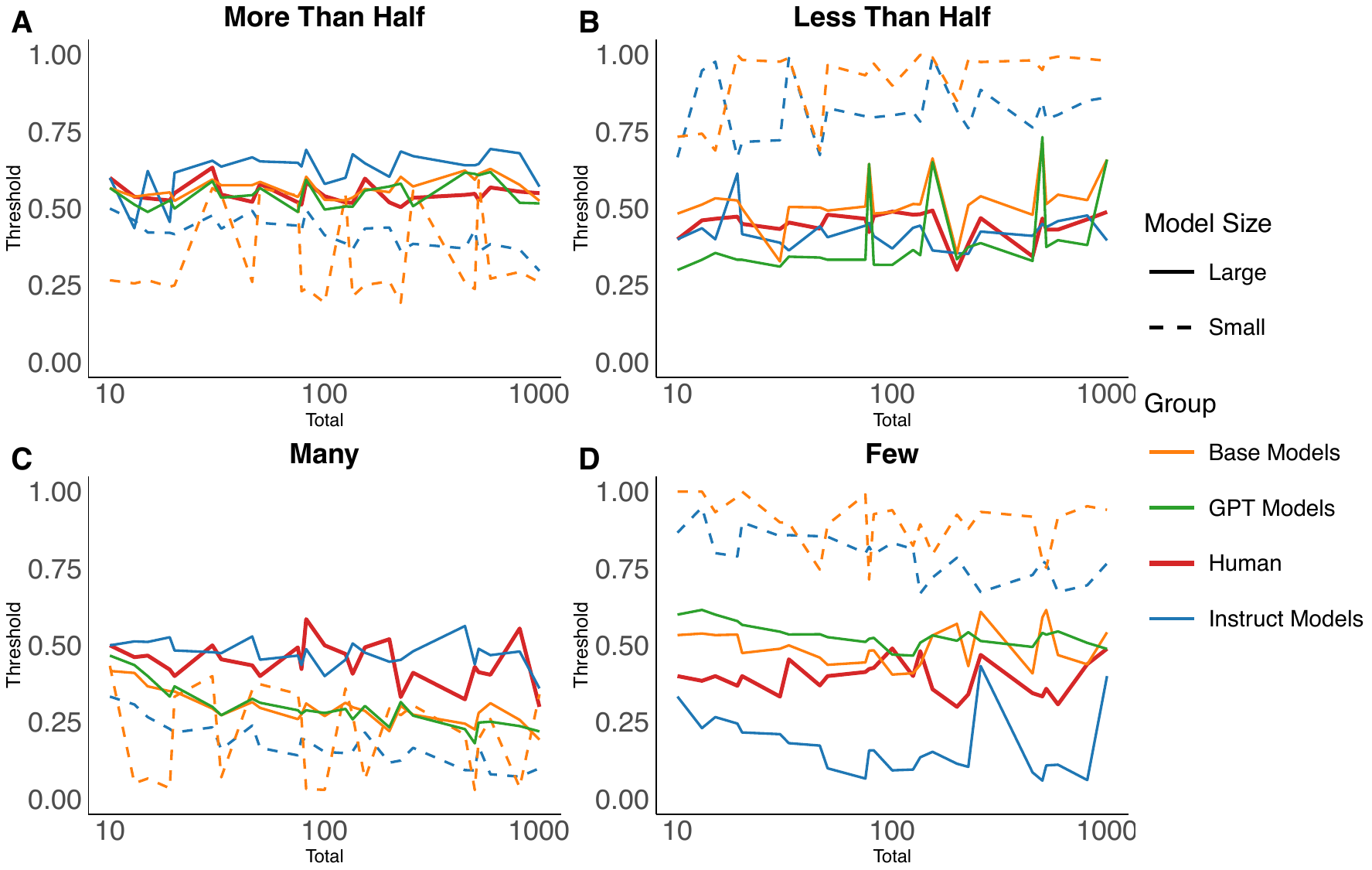}
\caption{Line plot with log(\texttt{<total-number>} on the x axis and threshold on the y axis. Dotted lines represent smaller models (< 10 billion parameters) and solid lines represent larger models (> 10 billion parameters). Human results are plotted in red as a bold solid line for reference. We compare vague vs exact quantifiers (with exact quantifier on the top and vague quantifiers on the bottom), and find that model performance on exact quantifiers (top row) are closer to human judgements than vague ones (bottom row).  We also compare results based on polarity (with positive quantifiers on the left and negative quantifiers on the right), and find that model responses for negative polarity quantifiers (right column) differ from human judgements more than positive polarity quantifiers(left column).}
\label{fig:line_plots_comparison}
\end{figure*}

The average difference between human and model judgments shown in Figure \ref{fig:avg_difference} exhibits a consistent trend that LLMs' performance on vague quantifiers is worse compared to exact quantifiers, the opposite of what we would expect for a model of distributional semantics. Only one model group (Small Base) does not demonstrate this trend. However, this group also has average threshold differences that are not significantly different from the maximum possible error (indicated by the dotted lines). Figure \ref{fig:line_plots_comparison} disaggregates these results and shows thresholds on each \verb|<total-number>|. In this figure, we also see closer alignment to human results on ``more than half'' (Panel A) compared to ``many'' (Panel C), and a closer alignment on ``less than half'' (Panel B) compared to ``few'' (Panel D).

\begin{table}[bt!]
\centering
\small
\caption{Regression slopes for all models and quantifiers. Boldface represents slopes that are statistically significantly different from 0 ($p$ < 0.05). ``IT'' indicates instruction-tuned. ``MH'' refers to ``more than half'', and ``LH'' refers to ``less than half''.}
\renewcommand{\arraystretch}{1}
\begin{tabular}{l|r|r|r|r}
\hline
Model & \multicolumn{1}{c|}{Many} & \multicolumn{1}{c|}{Few} & \multicolumn{1}{c|}{MH} & \multicolumn{1}{c}{LH} \\ 
\hline 
\hline
GPT-4 & \textbf{-0.103} & \textbf{-0.115}& -0.013 & 0.003 \\
GPT-4o & \textbf{-0.052} & -0.007 & \textbf{0.082} & \textbf{0.101} \\
GPT-3.5 & \textbf{-0.129} & 0.000 & 0.009 & 0.179 \\
LLaMA3-70b & -0.011 & 0.226 & 0.001 & 0.025 \\
LLaMA3-70b-IT & \textbf{-0.053} & \textbf{-0.112}& -0.000 & \textbf{0.070}\\
LLaMA3-8b & 0.028 & 0.000 & -0.034 & 0.000 \\
LLaMA3-8b-IT & \textbf{-0.166} & \textbf{-0.334} & \textbf{-0.168} & 0.004 \\
InternLM-7b & -0.022 & -0.049 & 0.026 & \textbf{0.272}\\
InternLM-7b-IT & \textbf{-0.075} & 0.000 &\textbf{0.070} & 0.000 \\
Qwen-72b & \textbf{-0.060} & 0.031 & 0.026 & 0.026 \\
Qwen-72b-IT & 0.031 & 0.046 & \textbf{0.036} & -0.017 \\
Qwen-32b & \textbf{-0.062}& \textbf{-0.115} & 0.010 & \textbf{-0.049} \\
Qwen-32b-IT & \textbf{-0.058} & \textbf{-0.079} & 0.103 & -0.067 \\
Mistral-7b & \textbf{-0.042} &0.083 & -0.005 & -0.003 \\
Mistral-7b-IT &\textbf{-0.058}& 0.059 & \textbf{-0.079} & \textbf{0.055} \\
Human & -0.033 & -0.003 & -0.009 & -0.008 \\
\hline
\end{tabular}
\label{tab:regression_slopes}
\end{table}

In addition, inspecting the responses for vague quantifiers versus exact quantifiers, we find qualitatively different patterns for LLMs compared to humans. While the threshold for exact quantifiers stays constant across different \verb|<total-number>|'s for language models, the threshold for vague quantifiers drops as \verb|<total-number>| rises.  To quantify this, we fit a linear regression separately for each model on $\log(\verb|<total-number>|)$ and present the slope and significance level in Table \ref{tab:regression_slopes}\footnote{We fit linear regression on $\log(x)$ instead of $x$ for more robust results. Appendix \ref{appendix:lm} explains this choice and includes linear regression fit on original data, where the same analyses still hold.}. We observe most of the significantly negative slopes are associated with vague quantifiers. However, the slopes associated with exact quantifiers are mostly either not significant (e.g GPT-4, GPT-3.5, Qwen-32b-IT), or are positive (e.g GPT-4o, Llama3-70b-IT). A negative slope suggests that the models are not abiding by strictly proportional readings of these quantifiers. One possibility that is consistent with the linear regression results is that models might be following a cardinal threshold, where the threshold value is fixed regardless of \verb|<total-number>|. While the exact processing that underlies language models warrants further explanation, this observation further strengthens the claim that models capture the semantics of exact quantifiers better than vague quantifiers.

\subsection*{Finding 2: Instruction-tuned models align better with human judgements}

\begin{table}[bt!]
\centering
\caption{Average difference with human between types of models. ``MH'' stands for more than half and ``LH'' stands for less than half. Bold text represents the largest difference score (worst performance). Base and Instruction Groups each include 6 models, and the GPT Group includes 3 models.}
\begin{tabular}{ l|c|c|c|c } 
\hline
  \multirow{2}{*}{Model Group} & \multicolumn{2}{c}{Positive} & \multicolumn{2}{c}{Negative} \\
 &Many& MH & Few& LH \\
\hline
\hline
Instruct & 0.161 & 0.309 & 0.104 & 0.209 \\ 
Base & \textbf{0.210} & \textbf{0.313} & \textbf{0.163} & \textbf{0.341} \\ 
GPT & 0.157 & 0.067& 0.277&  0.168 \\ 
\hline
\end{tabular}
\label{tab:model_type}
\end{table}

Table \ref{tab:model_type} shows the average difference to human judgements when models are grouped according to their type of training (e.g., instruction tuning vs.\ base). We see a consistent trend across all quantifiers: base models consistently have the worst alignment with human judgements, and GPTs consistently have the best. While the close human alignment of GPT models can be partially accounted by their overall performance, the instruct group and base group models consists of exactly the same models except for instruction tuning, but instruction-tuned models also consistently outperform base models.

We also see from Table \ref{tab:model_type} that instruction tuning has an overall higher effect on boosting human alignment on exact quantifiers than vague quantifiers. Given that post-training tends to focus on performance on tasks that include math and question answering, this result is somewhat intuitive. We discuss the implications of these results for treating instruction-tuned LLMs as distributional semantics models in Section \ref{sec:lm-dsm}.

\subsection*{Finding 3: Negative polarity quantifiers are harder to process}\mbox{}
Negative quantifiers (``less than half'' and ``few'') appear to be overall harder to process for models compared to their positive counterparts (``more than half'' and ``many''). From the horizontal comparison between the left and right panels of Figure \ref{fig:avg_difference}, the distance between models and human judgment is overall larger for negative quantifiers (right panel) versus their positive counterpart (left panel). Negative quantifiers also shows greater sensitivity in model size compared to positive quantifiers from Figure \ref{fig:line_plots_comparison}. Both instruct and base models in smaller sizes (shown in dotted lines) are distinctly distant from human judgement in negative quantifiers, contrary to the inverse scaling effect found in \cite{michaelov2023rarelyproblemlanguagemodels, gupta2023probingquantifiercomprehensionlarge}. 

Although further work is needed to diagnose exactly why this pattern arises, it is not inconsistent with observations from prior literature. First, humans exhibit a similar trend, both empirically in our experiments (slightly higher standard deviation for negative quantifiers in human results) and in previous research \cite{bott2013}. As language models are trained with datasets reflecting human language, their performance might thus imitate humans' ``weaker'' ability in dealing with negative quantifier. Second, it's not uncommon that language models struggle with negative statements. This has been well documented from the era of BERT-based models \cite{truong-etal-2022-improving,ettinger-2020-bert} through modern LLMs \cite{chen2023saymeanlargelanguage}. \citet{truong-etal-2023-language} also found a better performance on negation-related tasks from Instruction-tuned models, consistent with our observations. 

\section{Discussion \& Related Work}\label{sec:discussion}

\subsection{Are LLMs Models of Distributional Semantics?}
\label{sec:lm-dsm}
Conventional wisdom is that distributional semantics models (DSMs) excel at modeling semantic information that is inherently graded and based on use, but struggle on inferences that depend on compositional and truth-conditional reasoning. Our experiments show that, in a specific case involving graded (``many''/``few'') vs. exact (``more/less than half'') quantifiers, LLMs better track human judgments on exact inferences than on graded ones. It follows, then, that LLMs either are not distributional semantics models, or distributional semantics does not have the features it has been traditionally, sometimes definitionally, associated with.

We first entertain the argument that LLMs are not distributional semantics models. This argument seems difficult to make given seminal reviews of DSMs \cite{erk, boledaReview2020, lencireview}. Precise definitions of DSMs are hard to find, with most sources characterizing them vaguely, for example, as models in which “statistical distribution of linguistic items in context plays a key role in characterizing their semantic behavior”. These definitions are flexible enough to encompass many if not most computational models, depending on how one interprets “statistical”, “distribution”, and “key role”. A more useful way to define what has been meant by “DSM” historically is by example. The earliest DSMs were based on sparse, high-dimensional “bag of words” representations \cite{Turney_2010}, but these representations shortly after began to undergo dimensionality reduction, first through matrix factorization \cite{ir1990} and later neural networks \cite{cbow, NIPS2013_9aa42b31}. Models which were predictive (rather than count-based) have long been seen as DSMs \cite{baroni-etal-2014-dont}, and the possibility that distributional signal can include higher-order abstractions over surface statistics has been taken as given for decades \cite{landauer1997solution}. Models which update randomly initialized embeddings during training have been comfortably considered to be DSMs \cite{kanerva2000random}, as have models which include order and syntactic information \cite{lencireview, levy-goldberg-2014-dependency}, and context beyond text \cite{ lazaridou-etal-2015-combining}. 

Given this historical context, none of the obvious features which differentiate LLMs from their predecessors seem categorically different enough that LLMs should not be considered DSMs. Scale, for example, has never been a constitutive feature of DSMs; DSMs are not definitionally small, and making them bigger hardly seems to constitute a change in kind. The internals of model architecture --- i.e., the number of layers or the presence of attention --- seems a similarly arbitrary feature on which distributional semantics should hinge. Traditionally, the defining feature of DSMs lies in how they're trained; the internal or mathematical mechanism used to process this information is not explicitly defined nor constrained in any of the most influential reviews on the topic \cite{erk, boledaReview2020, lencireview}. A third possibility is that LLMs differ from prior DSMs because they use instruction tuning \cite{wei2022finetunedlanguagemodelszeroshot} and reinforcement learning from human feedback (RLHF) \cite{ouyang2022traininglanguagemodelsfollow}. This too is an unsatisfying argument. Instruction tuning amounts to changing the distribution of text on which the model is trained. Certainly a model cannot be disqualified as a DSM because of details of the distribution on which it is trained. If anything, evidence that the distribution alone has a substantive enough effect on the representations such that one distribution appears to exhibit sensitivity to truth-conditional meaning representations and another does not seems to further exacerbate the paradoxical situation. It might be possible to argue that RLHF --- that is, explicitly updating the weights of an LLM to maximize human preferences --- breaks some assumptions of distributional semantics, as it has been argued that RLHF models are not “text-only” models the way earlier LLMs are \cite{mahowald2024dissociatinglanguagethoughtlarge}. However, distributional semantics has never been limited to text only \cite{erk, lazaridou-etal-2015-combining}, and the earliest discussions of DSMs leave open the possibility of non-linguistic context, including social and goal-oriented “contexts”, being a part of the distribution. Thus, it is hard to see how the presence of RLHF alone makes LLMs no longer distributional.

One fair concern is that many state-of-the-art models are closed source, and there has been increased interest in incorporated neuro-symbolic components, or “tool use” \cite{parisi2022talmtoolaugmentedlanguage, cheng2023binding}, into LLM-based agents. Indeed, if it turned out LLMs were using calculators or Python interpreters under the hood in order to process exact quantifiers, that would disqualify them as DSMs. For the closed-source models, we cannot be sure this is not the case. However, the trend between between open source models and proprietary ones are the same, meaning that even if proprietary models are not DSMs, it does not change the core observations at issue in this paper.

\subsection{What can DSMs learn?}

Assuming LLMs are indeed still DSMs, then the question becomes about the fundamental nature of what DSMs can or cannot learn. It has long been assumed that DSMs should succeed in exactly the cases which require graded or subjective reasoning. The fact that our current best examples of DSMs seem to fair worse in such cases than in cases that require exact reasoning is a paradox and warrants explanation. Of course, there are many ways in which the distribution on which LLMs trains likely matters --- earlier arguments about distributional semantics might not have considered training on math and code, for example. Since the criteria for being a DSM give little to no constraint on the specific architectures and inductive biases of models, a possibility that remains open is that certain architectural biases could lead to a model's learning internal mechanisms that resembles formal, logical processing more than an associative model of co-occurrence \cite{doi:10.1177/09637214241268098, doi:10.1098/rsta.2022.0041}. Though there are evidence for transformers' learning logic-like symbolic operations under correct training conditions \cite{power2022grokkinggeneralizationoverfittingsmall, todd2023function}, such evidence are overshadowed by findings that LLMs are sensitive to distributional cues \cite{mccoyPNASambersofauto, ettinger-2020-bert, webson-pavlick-2022-prompt}. With LLMs still being treated like ``black boxes'', substantial further work is needed for a conclusive argument supporting either side. Point being, there are many avenues of investigation which could substantially refine previous arguments about what a DSM is and what it can do. And though these refinements are likely to be nuanced, subtle even, they are likely to be far from trivial. In fact, engaging with paradoxes like what we present here is likely to yield the type of theoretical breakthroughs for linguistics that have been boldly promised \cite{piantadosi2024modern} but lacking in practice. Distributional semantics has always been a realm in which computational and theoretical linguistics enhance one another. LLMs can and should carry on that tradition.

\section{Conclusion}
We investigate whether LLMs conform to predictions made from commonly-held assumptions about models of distributional semantics in the domain of quantifiers. Through comparing LLMs with human judgements, we find that the opposite of the prediction holds true: LLMs' performance on vague quantifiers is consistently worse than their performance on exact quantifiers. This contrasts with the claim that distributional semantics models should excel at modelling graded, vague meanings inferred from use but struggle with symbolic, logical inference. We then situate our findings in the broader historical and theoretical context of distributional semantics models (DSMs). Our results indicate that previous theories are insufficient for explaining the behaviors of LLMs, and refinement of such theories can be a fruitful path for both theoretical and computational linguistics.

\section{Limitations}

Designing empirical experiments for LLMs to test their alignment to predictions made by linguistic theories come with challenges. First, our study is limited to the four tested quantifiers, where a parallel between exact and vague quantifiers gives controlled settings for testing the distributional hypothesis, and does not test predictions made by distributional/formal semantics in more general linguistic domains. A fuller picture may be revealed through testing similar hypothesis on a larger variety of domains and contexts. Additionally, our experiment makes simplifications on the nuanced nature of quantifier processing: the extent to which context and individual linguistic environments affects judgements on quantifiers, especially vague quantifiers, are left largely unexplored. Lastly, our experiments are limited to behavioral evaluations of language models and comparing such results to humans. It presents a finding that warrants further explanations, but the research program of understanding linguistic processing in either LLMs or humans are still very much open questions this study does not directly tackle.

\bibliography{custom}

\appendix

\section{Experimental Design: Human} \label{appendix:exp_humans}
We started with randomly sampling 25 \verb|<total-number>| between 1 and 1000. Afterwards, we followed the sample criteria as described in Section \ref{sec:human_section} and obtained the \verb|<total-number>| \verb|<number>| pair in Table \ref{tab:total-pairs}. Note that the criteria do not guarantee 20 sample points for each of the total number. For instance, it's not possible to sample 20 distinct numbers under 10. Thus, some of the \verb|<number>| are repeated in the experiment. 

\begin{table}[ht!]
\small
\caption{Number pairs sampled and used in human experiments}
\begin{tabular}{c|l}
Total & Number samples \\
10 & 0, 1, 2, 3, 4, 5, 6, 7, 8, 9, 10 \\
13 & 1, 2, 3, 4, 5, 6, 7, 8, 9, 10, 11, 12, 13 \\
15 & 1, 2, 3, 4, 5, 6, 7, 8, 9, 10, 11, 12, 13, 14, 15 \\
19 & 1, 2, 3, 4, 5, 6, 7, 8, 9, 10, 11 \\
    & 12, 13, 14, 15, 16, 17, 18, 19 \\
20 & 1, 2, 3, 4, 5, 6, 7, 8, 9, 10, 11 \\
    & 12, 13, 14, 15, 16, 17, 18, 19 \\
30 & 1, 2, 4, 5, 7, 8, 9, 10, 13, 15, 19\\
    & 20, 21, 23, 25, 26, 27, 28, 29, 30 \\
33 & 2, 3, 5, 6, 7, 8, 9, 10, 11, 15, 18\\
    & 20, 21, 22, 23, 25, 28, 29, 30, 32 \\
46 & 1, 4, 5, 7, 10, 11, 15, 16, 17, 20, 24\\
    & 25, 26, 30, 32, 38, 40, 42, 43, 45 \\
50 & 3, 5, 10, 12, 15, 17, 20, 23, 24, 25, 29\\
    & 32, 34, 35, 40, 43, 45, 48, 50 \\
75 & 1, 2, 5, 15, 22, 25, 30, 31, 35, 37, 39\\
    &41, 45, 50, 51, 55, 57, 60, 73, 75 \\
78 & 6, 10, 12, 15, 19, 20, 25, 30, 31, 33\\
    &39, 40, 48, 49, 52, 55, 62, 65, 70, 75 \\
82 & 3, 4, 5, 9, 10, 15, 20, 27, 35, 39, 48\\
    & 50, 55, 60, 62, 63, 64, 66, 70, 75 \\
100 & 1, 8, 10, 24, 30, 31, 40, 45, 49, 50, 54\\
    & 55, 56, 60, 65, 67, 78, 80, 82, 85 \\
125 & 4, 6, 7, 8, 20, 30, 45, 50, 59, 60\\
    & 65, 70, 75, 86, 94, 101, 103, 115, 124, 125 \\
135 & 2, 8, 15, 23, 24, 45, 47, 55, 60, 65, 70\\
    & 78, 85, 89, 100, 105, 112, 113, 114, 135 \\
154 & 15, 18, 21, 23, 25, 35, 37, 55, 70, 76\\
    &79, 92, 95, 112, 115, 120, 129, 135, 149, 150 \\
200 & 11, 20, 23, 25, 26, 29, 45, 49, 55, 60, 104\\
    &105, 112, 115, 135, 136, 141, 145, 152, 155 \\
226 & 5, 25, 41, 55, 59, 65, 75, 77, 87, 114, 125\\
    & 126, 155, 185, 197, 200, 202, 214, 215 \\
258 & 22, 50, 51, 55, 65, 75, 85, 106, 117, 121, 138\\
    & 140, 161, 170, 188, 192, 197, 225, 235, 240 \\
450 & 10, 17, 26, 27, 50, 63, 100, 140, 146, 155, 245\\
    & 263, 270, 325, 340, 342, 364, 379, 415, 426 \\
500 & 22, 30, 96, 100, 115, 150, 167, 215, 227, 234, 274\\
    & 275, 313, 335, 365, 370, 406, 427, 430, 496 \\
521 & 10, 71, 90, 111, 112, 150, 187, 203, 215, 225 \\
    & 275, 285, 336, 343, 351, 380, 461, 475, 489, 500 \\
591 & 12, 25, 35, 100, 106, 128, 182, 239, 255, 270, 336 \\
    & 373, 380, 399, 425, 470, 495, 497, 565, 581 \\
809 & 5, 86, 140, 220, 234, 270, 284, 319, 355, 376, 449 \\
    & 450, 463, 540, 570, 708, 710, 732, 785, 793 \\
1000 & 100, 162, 165, 280, 283, 300, 400, 450, 489, 500 \\
     & 550, 614, 644, 650, 750, 767, 823, 900, 971 \\
\end{tabular}
\label{tab:total-pairs}
\end{table}

Table \ref{tab:noun-dis} demonstrates the selected nouns for each group, where group indicates the set of prompts each participant receives. 

\begin{table}
\small
\centering
\caption{Noun distribution across Groups. Each Group has 50 prompts with two nouns. }
\begin{tabular}{c|l}
\hline
    Group & Noun \\
    \hline
    0 & car, phone\\
    1 & ball, table\\
    2 & book, pen\\
    3 & bicycle, cup\\
    4 & phone, book\\
    5 & table, bicycle\\
    6 & computer, car\\
    7 & cup, computer\\
    8 & pen, clock\\
    9 & clock, ball\\
    10 & car, phone\\
    11 & ball, table\\
    12 & book, pen\\
    13 & bicycle, cup \\
    14 & phone, book\\
    15 & table, bicycle\\
    16 & computer, car\\
    17 & cup, computer\\
    18 & pen, clock\\
    19 & clock, ball\\
\hline 
\end{tabular}
\label{tab:noun-dis}
\end{table}

\begin{table}[bht!]
\caption{Sentences used in attention trials and practice trials.}
    \centering
    \small
    \begin{tabular}{c|l}
    \hline
        Type & Sentence \\
        \hline
        Attention &  There are 100 balloons.\\ 
        & 99 of them are yellow.\\
        & Are all of the balloons yellow?  \\
        \hline
        Attention &  There are 22 umbrellas.\\ 
        &  2 of them are yellow. Would you say \\
        &   at least one umbrella is yellow?   \\
        \hline
        Practice &  There are 10 bags. \\  
        & 1 of them is yellow. Would you say \\
        & one of the bags is yellow?   \\  
        \hline
        Practice & There are 100 water bottles.\\
        &99 of them are blue. Would you say \\
        &all of the water bottles are blue? \\
        \hline
        \end{tabular}
    \label{tab:practice-attention}
\end{table}
Recall that each trial of the experiment is formatted in this way: 
\vspace{0.5em}
\noindent

Question: There are \verb|<total-number>| \verb|<noun>|. \verb|<number>| of them are yellow. Are many \verb|<noun>| yellow? If yes, please press ‘J’ on your keyboard, otherwise press ‘F’.   

\vspace{0.5em}
For each experiment, there exist 20 groups (sets of stimulus) to be assigned to 20 participants. For each group, there are 50 sentences with 2 different \verb|<noun>| (as in \ref{tab:noun-dis}). The 25 sentences of the same noun corresponds to the 25 different \verb|<total-number>|'s paired with one of the \verb|<number>| (as in \ref{tab:total-pairs}). Thus, in order to cover the \verb|<number>| samples for a certain \verb|<total-number>|, it would require 20 participants (corresponding to the 20 groups) to take the experiment. 

Instructions include the purpose of data collection, as well as experiment procedures. Participants were told that they were participating in a research study, no personally identifying data is collected, and their data will only be used for the purpose of research studies. The experiment starts with instructions, then proceeds to the 2 practice trials. Then participants are asked to to answer the 50 trials and the 2 attention trials in a shuffled order. Prompts we used in attention and practice trials are presented in Table \ref{tab:practice-attention}.

We have in total 4 experiment for the 4 quantifiers we tested in the paper. Everything is kept constant except that the quantifier was substituted in each test trial. 

No personally identifying was collected. Participants received on average 10 dollars per experiment and the median time for each experiment was five minutes.

\section{Experimental Design: Language Models} \label{appendix:exp_lms}

To make results from language models maximally comparable to those from humans, we query the same set of questions given to humans to all lanugage models. Instead of splitting each set of experiments into groups of questions (as Section \ref{sec:human_section} describes), we give each LLM the all 1000 questions individually (with no shared contexts) and take the first output token. For all models, this output token often either ``\verb| Yes|'', ``\verb| No|'', or one of its equivalents (such as the lower case versions). For open-weight models, we run inference with weights of datatype \texttt{bfloat16}, and take the probability of all ``Yes'' tokens as well as all ``No'' tokens. The \emph{proportion} is then calculated as $\frac{P(\text{Yes})}{P(\text{Yes}) + P(\text{No})}$. Since proprietary models we used (from OpenAI) does not offer an API endpoint that allows querying for probabilities of arbitrary tokens, we approximate $\frac{P(\text{Yes})}{P(\text{Yes}) + P(\text{No})}$ by substituting $1 - P(\text{Yes})$ in place of $P(\text{No})$ when $P(\text{No})$ does not appear in query results as one of the top probability tokens, and vise versa.

Each language model receives the 1000 questions 5 times, each time with a different way of asking the question, but the same context. This helps mitigate prompt sensitivity of language models. Using ``many'' and ``balls'' to contextualize the examples, the 5 question templates were:

\begin{enumerate}
    \item \texttt{Are many balls yellow?}
    \item \texttt{Does it match the statement that many balls are yellow?}
    \item \texttt{Is it the case that many balls are yellow?}
    \item \texttt{Do you think less than half of the balls are yellow?}
    \item \texttt{Would you say many balls are yellow?}
\end{enumerate}

Each of the question templates were verified by multiple native speakers for acceptability. Thresholds are computed independently for each question template, and then averaged (while calculating standard deviation) for presenting in Section \ref{sec:results-lm}.

\section{Linear Regression Results on Original Data}
\label{appendix:lm}

Table \ref{tab:regression_original} presents the regression coefficients of all models, where the independent variable is \texttt{<total-number>} (instead of $\log(\texttt{<total-number>})$, as presented in Section \ref{sec:results-lm}). Given the distribution of \texttt{<total-number>}'s from 0 to 1000, one may reasonably suppose that performing a regression using $\log(\texttt{<total-number>})$ is going to be more robust, as high \texttt{<total-number>}'s are sparse, and doing regression on such data may lead to results being disproportionately affected by random noise in these data points. That being said, comparing Table \ref{tab:regression_original} to Table \ref{tab:regression_slopes}, one can see that the significant slopes and their trends remain the largely similar, both qualitatively and quantitatively: the analysis presented in Section \ref{sec:results-lm} still holds, and the significance of slopes is only different for 5 entries out of a total of 64 entries.

\begin{table}[bt!]
\centering
\small
\caption{Linear Regression slopes for all models and quantifiers. Boldface represents slopes that are statistically significantly different from 0 ($p$ < 0.05). ``IT'' indicates instruction-tuned. ``MH'' refers to ``more than half'', and ``LH'' refers to ``less than half''.}
\renewcommand{\arraystretch}{1}
\begin{tabular}{l|r|r|r|r}
\hline
Model & \multicolumn{1}{c|}{Many} & \multicolumn{1}{c|}{Few} & \multicolumn{1}{c|}{MH} & \multicolumn{1}{c}{LH} \\ 
\hline 
\hline
GPT-4 & \textbf{-0.0002} & \textbf{-0.0002} & -0.0000 & 0.0000 \\
GPT-4o & {-0.0001} & 0.0000 & \textbf{0.0001} & \textbf{0.0001} \\
GPT-3.5 & \textbf{-0.0002} & 0.0000 & 0.0000 & 0.0004 \\
LLaMA3-70b & -0.0000 & \textbf{0.0006} & -0.0000 & 0.0001 \\
LLaMA3-70b-IT & \textbf{-0.0001} & \textbf{-0.0002} & -0.0000 & \textbf{0.0001} \\
LLaMA3-8b & -0.0000 & 0.0000 & -0.0003 & 0.0000 \\
LLaMA3-8b-IT & \textbf{-0.0003} & \textbf{-0.0006} & \textbf{-0.0003} & 0.0000 \\
InternLM-7b & -0.0000 & -0.0000& 0.0001 & {0.0004} \\
InternLM-7b-IT & \textbf{-0.0001} & 0.0000 & \textbf{0.0001} & 0.0000 \\
Qwen-72b & {-0.0001} & 0.0000 & 0.0001 & 0.0000\\
Qwen-72b-IT & 0.0000 & 0.0003 & 0.0000 & 0.0000 \\
Qwen-32b & \textbf{-0.0001} & \textbf{-0.0002} & 0.0000 & \textbf{-0.0001} \\
Qwen-32b-IT & {-0.0001} & {-0.0001} & 0.0001 & -0.0001 \\
Mistral-7b & -0.0001 & -0.0001 & 0.0000 & 0.0000 \\
Mistral-7b-IT & \textbf{-0.0001} & 0.0001 & \textbf{-0.0002} & \textbf{0.0001} \\
Human & -0.0001 & 0.0000 & -0.0000 & -0.0000 \\
\hline
\end{tabular}
\label{tab:regression_original}
\end{table}

\end{document}